\documentclass[11pt,a4paper]{article}
\usepackage[utf8]{inputenc}
\usepackage[english]{babel}
\usepackage{amsmath}
\usepackage{amsfonts}
\usepackage{amssymb}
\usepackage{makeidx}
\usepackage{graphicx}
\usepackage{lmodern}
\usepackage{bm}
\usepackage{multirow}
\usepackage{supertabular}
\usepackage{dcolumn}
\usepackage{pdflscape}
\usepackage{pgfplots}
\usepackage{accents}
\usepackage{setspace}
\usepackage[round]{natbib}
\usepackage{lscape}
\usepackage[left=2cm,right=2cm,top=2cm,bottom=2cm]{geometry}

\begin{document}

{\center {\Large Deep learning for detecting bid rigging:\\Flagging cartel participants based on convolutional neural networks}{\large
\vspace{0.1cm}}\smallskip\\

{\large Martin Huber* and David Imhof**}\smallskip\\
{\small {* University of Fribourg, Dept.\ of Economics }}\\[0pt]
{\small {** Corresponding Author, Swiss Competition Commission, University of Fribourg, Dept.\ of Economics, and Unidistance (Switzerland)}}\\[0pt]}\smallskip

\vspace{2cm} \noindent \textbf{Abstract:} {\small \textit{Adding to the literature on the data-driven detection of bid-rigging cartels, we propose a novel approach based on deep learning (a subfield of artificial intelligence) that flags cartel participants based on their pairwise bidding interactions with other firms. More concisely, we combine a so-called convolutional neural network for image recognition with graphs that in a pairwise manner plot the normalized bid values of some reference firm against the normalized bids of any other firms participating in the same tenders as the reference firm. Based on Japanese and Swiss procurement data, we construct such graphs for both collusive and competitive episodes (i.e when a bid-rigging cartel is or is not active) and use a subset of graphs to train the neural network such that it learns distinguishing collusive from competitive bidding patterns. We use the remaining graphs to test the neural network's out-of-sample performance in correctly classifying collusive and competitive bidding interactions. We obtain a very decent average accuracy of around 90\% or slightly higher when either applying the method within Japanese, Swiss, or mixed data (in which Swiss and Japanese graphs are pooled). When using data from one country for training to test the trained model's performance in the other country (i.e.\ transnationally), predictive performance decreases (likely due to institutional differences in procurement procedures across countries), but often remains satisfactorily high. All in all, the generally quite high accuracy of the convolutional neural network despite being trained in a rather small sample of a few 100 graphs points to a large potential of deep learning approaches for flagging and fighting bid-rigging cartels.}}
\vspace{0cm}\smallskip\\

{\footnotesize \noindent \textbf{Keywords:} Bid rigging, deep learning, convolutional neuronal networks, bid rotation test.}\vspace{2pt}

{\footnotesize \noindent \textbf{JEL classification:} C21, C45, C52, D22, D40, K40, L40, L41.}\vspace{2pt}

{\footnotesize \noindent \textbf{Address for correspondence:} David Imhof, Hallwylstrasse 4, 3003 Bern, Switzerland; david.imhof5@gmail.com.}\vspace{2pt}

%{\footnotesize \noindent \textbf{Acknowledgement} The authors would like to thank *** for support and helpful comments.}

{\footnotesize \noindent \textbf{Disclaimer:} All views contained in this paper are solely those of the authors and cannot be attributed to the Swiss Competition Commission, its Secretariat, the University of Fribourg, Unidistance (Switzerland).}

\thispagestyle{empty}\pagebreak  {\small
\renewcommand{\thefootnote}{\arabic{footnote}}
\setcounter{footnote}{0} \pagebreak \setcounter{footnote}{0}
\pagebreak \setcounter{page}{1} }

\newpage

\section{Introduction}

Bid rigging or collusive tendering is a pervasive problem in many markets and countries \citep[see for instance]{Feinstein1985, Porter1993, Baldwin1997, Porter1999, Pesendorfer2000, Banerji2004, Abrantes2006, Lee2002, Bajari2003, Asker2010, Ishii2009, Abrantes2012, Ishii2014, Hueschelrath2014, Bergman2020} and damages the taxpayer through substantially increasing the cost of procurement in public tenders. The OECD estimates that the elimination of bid rigging might reduce procurement prices in tenders by 20\% or more\footnote{See http://www.oecd.org/competition/cartels/fightingbidrigginginpublicprocurement.htm, accessed in March 2021.} and recommends developing pro-active tools for detecting bid-rigging cartels to supplement the traditional means as the leniency programs \citep[see][]{OECD2014}. Indeed, a growing literature  proposes statistical tools for unveiling cartels \citep[see][among others papers]{Abrantes2006, Bolotova2008, Harrington2007, Abrantes2012, Imhof2018, Imhof2019, Chassang2020}. Among them is an increasing number of studies applying machine learning for flagging suspicious tenders \citep[see][]{Foremny2018, Huberimhof2019, Rabuzin2019, Wallimann2020, Garcia2020,  Silveira2021}, which algorithmically learns to optimally predict cartels from historical data. In this paper, we add to this growing arsenal of artificial intelligence-based methods by proposing a novel approach for detecting bid-rigging cartels based on deep learning, which is an extension to machine learning that has revolutionized the quality of image recognition, natural language processing, and many other tasks.

We apply so-called convolutional neural networks (CNN) - a deep learning method never implemented before in the screening literature - to graphical plots of bid interactions of one firm with all other firms across various tenders in order to learn whether such interactions occur most likely under collusion or competition. Similar to conventional machine learning approaches, CNNs aim at developing optimal predictive models by learning from systematic patterns in data. In contrast to machine learning, however, they do not require predefining specific variables to be used as predictors in the model, but learn autonomously e.g.\ from graphs which features (e.g.\ specific shapes formed by pixels in images) are most relevant for recognizing an image. Due to their high performance in such context, CNNs appear to be a natural choice for developing methods for cartel prediction based on graphs of bidding interactions, permitting competition agencies to screen suspicious bids for deeper investigation and to disclose illegal collusive activities, with potentially large gains for tax payers.

We consider procurement data from Switzerland \citep[see][]{ Wallimann2020} and Japan \citep[see][]{Huberimhof2020}, in which we can distinguish collusive from competitive tenders, for training (i.e. developing) the CNN models for predicting bid-rigging cartels as well as for testing, i.e.\ assessing model performance in terms of prediction accuracy. As model inputs, we use graphs from the bid rotation test suggested by citet{Imhof2018} and \citet{Imhof2019}, which consists of depicting the strategic interaction of firms in the bidding process. To this end, the bids (i.e. prices) of one specific firm are plotted against the bids of several other firms participating in the same tenders within a quadrant that is normalized such that all bids contain values between zero and one. The prevalence of systematic patterns in such plots of pairwise bids permits detecting strategic interactions between bidding firms across tenders. Similar as in the image recognition of photos (e.g. classifying animals as cats or dogs), CNNs can learn from such bid plots to classify specific interactions as likely collusive or competitive even without prior knowledge about the nature of patterns underlying bid rigging. %The aim of this paper is thus to use a large number of plots as “images” to feed into CNN models in order to learn systematic patterns that are typical for collusion or competition. Using random subsets of the plots as training and testing samples, respectively, we investigate the correct prediction rate of competitive and collusive bids of such a method. Building an effective detection method based on CNNs permits competition agencies to screen suspicious bids for deeper investigation in order to disclose and terminate illegal collusive activities.

When assessing the model performance in the Swiss and Japanese data, we obtain an impressive overall accuracy (or correct classification rate) in classifying collusion and competition of roughly 91\% and 90\%, respectively. The results therefore suggest that our detection method based on CNN models correctly classifies nine firms out of ten as cartel participants or competitors. When pooling the graphs from both Switzerland and Japan to generate a transnational dataset of tenders, we even obtain a marginally higher accuracy of 91-92\%.
%Martin: why the result increases indeed? Is that because of the larger sample by combining Japan and Swiss?
We also consider training the CNN in data from one country and using the trained model for predicting collusion in the other country in order to test performance in an institutional context that is different to the training environment. When training the CNN based on Japanese graphs, we obtain an accuracy of 85-86\% when predicting collusion in the Swiss data. The performance is, however, somewhat lower when training based on Swiss bidding plots data and testing in the Japanese data, with accuracy amounting to 79\%.

We also observe that when training in one country to test in the other one, there are noticeable imbalances across the true positive and true negative rates, i.e.\ the correct classification rates among truly collusive or truly competitive firms. While Japanese competitive and Swiss collusive bid interactions are mostly correctly classified with an accuracy above 95\%, the performance is lower for Japanese collusive  and Swiss competitive interactions. This result suggests that the bidding patterns in some of the Japanese collusive and Swiss competitive graphs are insufficiently dissimilar for obtaining a high out-of-sample accuracy. Such difficulties in directly transferring trained models from one country to another might be related to institutional differences in the procurement process (e.g.\ the number of bids typically submitted), as also acknowledged by \citet{Huberimhof2020} when analyzing the Swiss and Japanese data based on a machine learning approach. However,  despite such imbalance, the overall correct classification rates of 85-86\% and 79\% appear quite decent given that the training and testing data come from distinct countries.
Moreover, the high accuracy obtained when training and testing within a single country (or even a mixed data set of several countries) suggests that the application of CNNs to graphs depicting the bidding interaction of firms provides national competition agencies with a very powerful tool for screening procurement markets.

%Martin: below the literature: I put sufficient materials to be complete. Do not hesitate to cut if it needs to be. Note however that it is always important to explain why the structural approach based on costs is difficult to apply because one needs data on the firm level (which are difficult to obtain without attracting the attention of the cartel).
Our paper is related to an expanding literature using screening methods for detecting cartels \citep[see][]{Abrantes2006, Harrington2007, Bolotova2008, Abrantes2012, Jimenez2012, OECD2014, Froeb2014}. More specifically, it considers statistical screens for flagging bid-rigging cartels as \citet{Imhof2018} and \citet{Imhof2019}. \citet{Huberimhof2019} use such screens (like the coefficient of variation or the kurtosis of the bids in a tender) as inputs to machine learning algorithms for building predictive models for collusion based on Swiss data. \citet{Wallimann2020} focus on the machine learning-based prediction of  incomplete (or partial) bid-rigging cartels by calculating a large number of screens based on sub-groups of bids in tenders. \citet{Huberimhof2020} apply machine learning techniques to a Japanese bid-rigging cartel and also investigate the transnational predictive performance across Swiss and Japanese data (as we also do for deep learning in this paper). All these studies focus on a tender-based approach for flagging bid-rigging cartels, i.e.\ they analyze the bid distributions within tenders. In this paper, we, however, focus on a firm-based approach that analyzes the bid interactions of firms in a specific region and period across tenders to test if their interactions are of collusive nature. Finally, \citet{Chassang2020} investigate the occurrence of “gaps” in bid values (i.e.\ bid values with zero probabilities) in the distribution of observed bids. Finding such gaps, especially between the first and the second lowest bid in a tender, might point to collusive behavior as they should not systematically occur in competitive markets. Related to our paper, this phenomenon of missing density implies that some areas of our plots remain empty when firms collude, which may be recognized as predictive feature when training a CNN for detecting collusion.

Our paper is also related to further studies on detecting bid-rigging cartels which apply econometric tests, as suggested in the seminal paper of \citet{Bajari2003}. A first test consists of verifying if bids are independent conditional on the costs of each firm by estimating a bidding function and testing if the residuals are correlated between firms, indicating potential collusive issues. A second test examines if firms react similarly as a function of their respective costs, based on the estimated cost-related coefficients in a regression model. Divergent coefficients across firms could indicate potential collusive issues. Subsequent papers replicate and refine the two econometric tests suggested by \citet{Bajari2003} \citep[see][]{Jakobsson2007, Aryal2013, Chotibhongs2012a, Chotibhongs2012b, Decarolis2016, Imhof2017b, Bergman2020}. However, when \citet{Imhof2017b} applies the econometric tests of \citet{Bajari2003} to the Ticino bid-rigging cartel, the tests fail to detect pairs of firms that actually colluded. As a further caveat, such econometric tests require data on the costs of each firm, which are typically not observed in bidding processes. It appears generally difficult to access data on the individual costs of firms, in particular when a competition agency wishes to apply ex ante tests for detecting bid-rigging cartels. Questionable predictive power and lack of information therefore likely impede the usefulness of econometric tests as a detection method for competition agencies such that machine or deep learning-based approaches for analyzing the bid distribution appear more promising. Finally, our paper is also broadly associated with further studies investigating bid-rigging cartels or rings \citep[see][]{Baldwin1997, Porter1999, Banerji2004, Lee2002, Asker2010, Ishii2009, Ishii2014}.

The remainder of this study is organized as follows. Section \ref{bidrotscreen} introduces the bid rotation screen that underlies our analysis and illustrates how we compute graphs for the bidding interactions of firms. Section \ref{data} introduces the Japanese and Swiss data considered in the empirical analysis. Section \ref{cnn} provides a brief introduction to deep learning based on CNNs and outlines our CNN architecture. Section \ref{empsec} presents the empirical results. Section \ref{con} concludes.

\section{The bid rotation screen}\label{bidrotscreen}

The inputs for predicting bid-rigging cartels in our empirical analysis consist of graphs depicting the bidding interaction between one firm and a group of other firms in a series of tenders during a specific period. We follow the ideas in \citet{Imhof2018} and \citet{Imhof2019} and use the so-called bid rotation screen to construct the graphs entering our algorithm. In order to plot the interaction of firms for several tenders of different contract values, we apply the so-called min-max transformation to some bid $i$ within some tender $t$ as follows:
\begin{equation}\label{normalisation}
\hat{b}_{it}=\frac{b_{it}-b_{min, t}}{b_{max, t}-b_{min, t}},
\end{equation}
where $b_{min, t}$ and $b_{max, t}$ are the minimum and maximum bids in tender $t$, respectively. Therefore, each transformed bid takes values between zero and one across all tenders, $\hat{b}_{it}   \in [0, 1]$. In a next step, we calculate the Cartesian coordinates of transformed bids for each possible pair of firms, i.e.\ of some reference firm and each of the other firms participating in the same tenders as the reference firm in a specific period, on the space $[0,1]\times[0,1]$. We then plot the Cartesian coordinates of these pairs of firms in a graph, thus depicting the interactions of the reference firm (on the x-axis) with a group of firms (on the y-axis) bidding in the same tenders as the reference firm.

We construct the graphs separately within collusive periods (with a cartel in place) and competitive periods (after a cartel has been dissolved) for each reference firm, in order to depict the firm's bidding interactions both within cartels and under (post-cartel) competition. Our suggested method exploits systematic differences in the bidding interaction under collusion and competition, but remains agnostic about how these differences look like. It simply relies on the plausible hypothesis that the bidding behavior of firms differs between the cartel and post-cartel periods in some way due to strategic behavior and aims at detecting such differences algorithmically by autonomously learning from data. This in turn implies that if the bidding behavior of firms was identical across cartel and post-cartel periods, our method (or any other approach based on screens or econometric tests) would not be able to detect bid-rigging conspiracies, but our empirical findings discussed further below suggest that this is not the case.

\begin{figure}
\caption{Example of non-competitive interactions between two firms}\label{graphexample}
\begin{center}
\includegraphics[height=12cm, width=12cm]{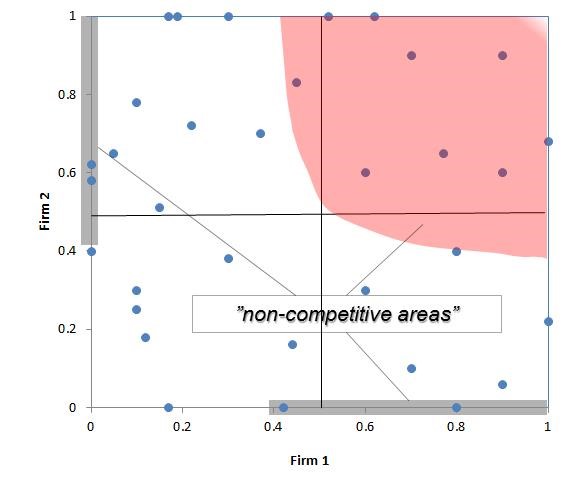}
\end{center}
\end{figure}

Even if the existence of some (and possibly unknown) kind of difference in distributional patterns of bids across collusive and competitive periods is sufficient for applying our method, several studies have suggested explanations for how bid rigging changes interactions, e.g. by inducing asymmetry in the distribution of bids \citep[see][for examples in Switzerland]{Huberimhof2019, Imhof2019} and \citep[see][for examples in Japan]{Chassang2020, Huberimhof2020}. In Swiss bid-rigging cases, for instance, differences between the two lowest bids in a tender increase whereas the differences between losing bids simultaneously decrease. The two effects combined create asymmetry in the distribution of bids, which can be captured by some screens applied on a tender basis. Also in Japanese data, bid rigging entails an asymmetry in the distribution of the bids, as revealed by some screens considered in \citet{Huberimhof2020}. Furthermore, \citet{Chassang2020} formalized a model implying that increasing differences between the first and second lowest bids are not in line with competition. Such a zero density in bid values between the first and second lowest bids renders the distribution of bids asymmetric. In a context of our min-max transformation, asymmetry in the distribution of bids (or zero density) implies that collusive bids are concentrated in specific regions of the two dimensional space $[0,1]\times[0,1]$, whereas competitive bids are scattered all the space. Considering the interaction of just two firms across different tenders, Figure \ref{graphexample} taken from \citet{Imhof2018} and \citet{Imhof2019} highlights regions (non-competitive areas) in the two dimensional space in which we would suspect a more concentrated density of bids in the case of bid rigging-induced asymmetry. This is either due to one firm submitting a comparably high bid to guarantee that the other, lower-bidding firm wins the contract, or due to both firms submitting high bids to allocate the contract to another cartel member (not depicted in the graph). Such distributional patterns would therefore indicate strategic interactions between firms and therefore point to collusive behavior.

%Martin: I will maybe change the following graphic. I had a better idea but just at the end.

\section{Japanese and Swiss bid-rigging data}\label{data}

Our empirical analysis is based on bidding data from Japan, namely the Okinawa region, and Switzerland. Concerning the Okinawa bid-rigging cartel, we refer to \citet{Ishii2014} and \citet{Huberimhof2020} for further details concerning the institutional context and data construction and consider all firms that submitted at least ten bids in the data window for our deep learning approach. %in the data including the pre-inspection, the post-inspection and the post-amendment period.
We focus on contracts of so-called ranks A+ and A in the Japanese data, with ranks depending on the reserve prices of the contracts as discussed \citet{Huberimhof2020}. The latter study found that the impact of the bid-rigging investigation launched by the Japanese Federal Trade Commission (JFTC) in June 2005  and the sanctions imposed on cartel participants in March 2006 were strongest for those ranks of contracts, in which condemned bidders had participated more frequently. This implies that for such contract types, it was particularly unlikely that former cartel participants immediately colluded again in post-cartel periods after sanctioning, which thus serve as competitive observations in our data.
	
%Moreover, \citet{Huberimhof2020} focused on a tender-based approach for screening markets and they used in their sample all tenders in which a condemned bidder participated. Following our interaction-based approach, we did not consider the contracts of rank B and C since condemned bidders participated in only some contracts of those types. We would not have been able to calculate the interaction of only several condemned bidders in those types of contracts. Finally, \citet{Huberimhof2020} also found that not all tenders for contracts of rank B and C were classified as competitive in the post-amendment period. To avoid the contamination of the sample (competitive graphics mixed with collusive ones) and the associated risk of lessening the predictive performance of trained models, we therefore focus on contracts A+ and A.

As in \citet{Huberimhof2020}, the pre-investigation period  from April 2003 to June 2005 presumably comprises the collusive tenders and the post-amendment period from January 2006 to May 2007 (after sanctioning) the competitive cases.\footnote{We do not consider contracts of so-called ranks B and C in which condemned cartel members participated less frequently such that the number of interactions for constructing the graphs would have been too low. Furthermore, findings in \citet{Huberimhof2020} suggest that there might be collusive interaction for these kind of contracts even after the sanctions by the JFTC such that including these post-amendment observations could contaminate our sample of competitive interactions and thus, reduce the predictive performance of our method.}  Concerning rank A+ contracts, we consider 48 tenders with 670 bids in the pre-investigation period and 57 tenders with 1049 bids in the post-amendment period. For contracts of rank A, we consider 108 contracts with 1325 bids in the pre-investigation period  and 86 contracts with 1439 bids in the post-amendment period. All in all, 161 firms bade for A+ contracts, 491 firms for A contracts, see Table \ref{bidrot1}.

{\renewcommand{\arraystretch}{1.1}
\begin{table} [!htp]
\caption{Japanese sample} \label{bidrot1}
\begin{center}
\begin{tabular}{llccc}\hline\hline
Type of contract&Period&Number of bids&Number of firms&Number of contracts\\
Rank A+&Collusive&1049&161&57\\
Rank A+&Competitive&670&159&48\\
Rank A&Collusive&1339&491&108\\
Rank A&Competitive&1439&481&86\\\hline\hline
\end{tabular}
\end{center}
\par
%{\footnotesize Note: “Num. of bids", “Num. of firms" and “Num. of contracts" denote the number of bids, the number of firms, and the number of contracts for each period and contract type, respectively.}
\end{table}}

\begin{figure}
\caption{Interactions of a Japanese firm with other bidders (left: cartel period; right: competitive period)}\label{graphjapfirm1}
\begin{center}
\includegraphics[height=8cm, width=12cm]{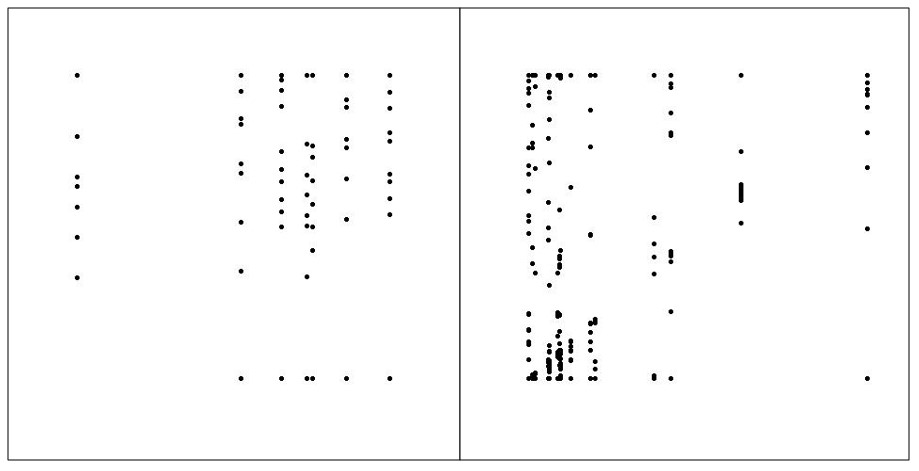}
\end{center}
\end{figure}

We only calculate interaction graphs for those reference firms submitting at least 10 bids for contracts of rank A+ and A during the data window, but consider all of the submitted bids in the relevant tenders to compute pairwise bidding interactions. This implies that even a firm who only participated in one tender in which the reference firm was present is considered in one of the pairwise interactions in the graph of that tender. %We then order the firms so that firm 1 is the firm who bids the most in the whole period.
As an example, Figure \ref{graphjapfirm1} provides the collusive and competitive graphs for contracts A+ computed for that firm that placed the most bids in our data window. The left graph depicts the strategic interaction of that firm during the pre-investigation period when the cartel was active. As expected, we find the bottom-left area to be empty, while we observe more pairwise interactions in the top-right region (likely supporting a cartel member designated to win the contract), in the upper area close to the y-axis, or in the right area close to the x-axis (such that one firm likely supports the other in that pair). The right graph shows the interactions of the reference firm during the post-amendment period and in line with competition, pairwise interactions also enter the lower-left area. All in all, we obtain 143 collusive and 144 competitive graphs from the Japanese data.

%\subsection{Graphs from the Swiss bid-rigging cartel}

Concerning Switzerland, we use data from the canton of Ticino that consist of both collusive and competitive periods \citep[see][]{Imhof2019} and observations from two other regions as the competitive periods \citep[see][]{Wallimann2020}. Compared to the Japanese data, the data window is much longer for the Ticino data, ranging from January 1999 to March 2005 for the cartel period. As reported in Table \ref{bidrot2}, we consider 1331 bids in 183 contracts for the collusive period. However, the number of bidding firms is considerably lower than in the Japanese data, as we only observe 21 cartel participants, see Table \ref{bidrot2}. We compute the graphs on a yearly base for those 16 firms who participate most in the tenders. In the much shorter competitive period from April 2005 to the end of 2006, we can exploit 297 bids in 40 tenders and compute graphs for the same 16 firms over that period (i.e.\ without dividing the competitive tenders into subperiods).

{\renewcommand{\arraystretch}{1.1}
	\begin{table} [!htp]
		\caption{Swiss sample} \label{bidrot2}
		\begin{center}
			\begin{tabular}{llccc}\hline\hline
				Data&Period&Number of bids&Number of firms&Number of contracts\\
				Ticino&Collusive&1331&21&183\\
				Ticino&Competitive&297&21&40\\
				Other Swiss regions&Competitive&3728&188&1018\\
				\hline\hline
			\end{tabular}
		\end{center}
		\par
		%{\footnotesize Note: “Num. of bids", “Num. of firms" and “Num. of contracts" denote the number of bids, the number of firms, and the number of contracts for each period, respectively.}
\end{table}}

\begin{figure}
		\caption{Interactions of a Swiss firm with other bidders (left: cartel period; right: competitive period)}\label{graphswissfirm1}
	\begin{center}
		\includegraphics[height=8cm, width=12cm]{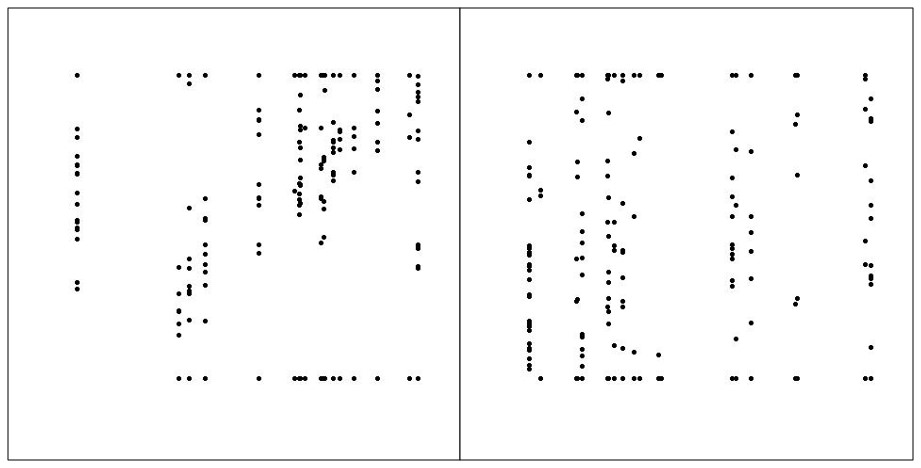}
	\end{center}
\end{figure}

Figure \ref{graphswissfirm1} reports bidding interactions of that firm that bade most frequently in the Ticino data. The left graph depicts the first year of the collusive period in 1999. Similar as for the Japanese data, the bottom-left area is empty in contrast to the top-right region. In the right graph, however, which provides the interactions in the competitive period, interactions occur in the bottom-left part, too, such that no apparent gaps in the values of pairwise bids of two firms occur. Figure \ref{graphswissfirm1} therefore suggests that the reference firm behaved differently in the year 1999 when compared to the competitive period.

Since the number of graphs is unsatisfactory and our Ticino sample is imbalanced with many more collusive than competitive graphs, we also consider data from competitive periods in other Swiss regions to obtain a more balanced sample \citep[see][for details on the cases]{Wallimann2020}. We to this end create graphs on a yearly base that depict the interactions of former cartel participants in competitive periods (after the cartel has been dismantled), based on 3728 bids in 1018 tenders. Combining these data with the Ticino case yields all in all 144 graphs for competitive periods and 96 for collusive periods in various Swiss regions.

\section{A convolutional neural network for detecting collusive firms}\label{cnn}

Convolutional neural networks (CNNs), as for instance discussed in \citet{LeCunetal1998} and \citet{Ciresanetal2011} are a powerful and increasingly popular deep learning approach for image (e.g.\ facial) recognition. By their nature, they can in principle learn any kind of patterns in plots of bidding interactions that are predictive for bid rigging from a large enough training data set with collusive and competitive cases, even if the shape such patterns might take is a priori unknown to the researcher. This distinguishes such a deep learning approach from previously applied machine learning methods that require so-called feature engineering, namely pre-specifying the features to be used as predictors, such as specific statistical screens like the coefficient of variation or the kurtosis. However, CNNs require choosing a range of model tuning (or hyper-)parameters that determine how the model extracts and learns the most predictive patterns and it needs to be pointed out that the choice of these parameters may crucially affect the performance of the method.

CNNs can be regarded as an extension of standard feed-forward neural networks (FNN), see e.g.\ \citet{McCulloch1943} and \citet{Ripley1996}. In the context of bid rigging, such FNNs aim at fitting a system of nonlinear functions that flexibly models the statistical association of a range of predictors (like statistical screens) and an outcome of interest (like a binary indicator for collusion). Specifically, the predictors serve as inputs for specific nonlinear intermediate regression functions (e.g.\ sigmoid or rectifier functions), so called hidden nodes, which themselves serve as inputs for modeling the outcome of interest. The hidden nodes bear some similarity with the baseline functions in spline or series regression, with the difference that they are learned from the data rather than predetermined, and with principal component analysis, which are dimension-reducing linear (rather than nonlinear) functions of the predictors. Indeed, when replacing the nonlinear functions by linear ones, FNNs collapse to a linear regression model. Depending on the model complexity, hidden nodes may affect the outcome either directly or through other hidden nodes, such that several layers of hidden nodes allow modeling interactions between the functions. The number of hidden nodes and layers gauges the flexibility of the model, with more parameters reducing the bias but increasing the variance.

In contrast to FNNs, CNNs do not rely on providing a list of predictors but may autonomously learn to derive relevant features from objects like images that do a priori not have a clear data structure. To this end, so-called filters (or kernels) are applied, which slide over a pre-specified amount of pixels in images to map them into numeric values based on nonlinear regression functions, which permits representing specific spatial patterns (such as edges) by numeric features. This is typically followed by a so-called pooling step that aggregates these features, e.g.\ by taking average or maximum values over a pre-specified number of adjacent numeric features. This may be followed by further filtering (of the pooled features) and pooling steps for transforming and aggregating the features. Finally, the refined features are used as inputs in a conventional FNN for prediction.

While CNNs do not require direct feature engineering as FNNs or classical machine learning, they obviously rely on making a range of design choices concerning how the important features should be learned by (possibly nested) filtering and pooling. In addition, one is confronted with the same design choices in FNNs concerning the number of hidden nodes and layers as well as further tuning parameters. The latter include the share of the training data, i.e.\ the part of the total sample used to develop a predictive model, and the validation data for assessing its predictive performance based on the accuracy (or correct classification rate) outside the training sample during the process of refining the model. A further tuning parameters is the batch size, the number of observations in the training data to work through before updating the model parameters (in particular the coefficients in the non-linear models) when aiming at minimizing the classification error in the training data based on so-called stochastic gradient descent. Finally, one needs to specify the number of epochs, i.e.\ of times that the algorithm works through the entire training data for estimating the model parameters.

Due to this large amount of choices to make that may apparently come with little guidance, tuning deep learning models is frequently regarded to be more an art than a science. Also in our empirical application, we tried some 10 to 15 different specifications in terms of numbers of hidden layers and nodes, filtering and pooling steps, as well as batch size and epochs. In the empirical section, we only present the results for the CNN specification outlined in Figure \ref{cnnarchitecture}, which performed better or at least not worse than any of the other specifications considered. Using this model, we set the share of training and validation data per epoch to 90\% and 10\%, respectively,  the batch size to 16, and the number of epochs to 40. Other than this we use the default options of the “keras" package by \citet{Falbeletal2021} for the statistical software “R", an interface to  the “Keras"  application programming interface (API) for neural networks, which itself makes use of the open source software library “TensorFlow" for implementing deep learning architectures.

\begin{figure}
	\caption{CNN architecture for predicting collusion}\label{cnnarchitecture}
\begin{enumerate}
	\item Convolutional layer (two-dimensional for gray-shaded images) with 8 filters, each with kernel size 3$\times$3, using the rectifier (or ReLU) as nonlinear function for mapping pixels into numeric values.
	\item Max pooling with size  2$\times$2.
	\item Convolutional layer with 16 filters, each with kernel size 3$\times$3, using the rectifier as nonlinear function for mapping the pooled features.
	\item Max pooling with size  2$\times$2.
	\item Convolutional layer with 32 filters, each with kernel size 3$\times$3, using the rectifier as nonlinear function for mapping the pooled features.
	\item Flattening: preparing the features generated by the previous convolutional layer as input variables for FNN to follow.
	\item Hidden layer with 128 hidden nodes, using the rectifier as nonlinear function for mapping the inputs.
	\item Hidden layer with 64 hidden nodes, using the rectifier  as nonlinear function for mapping the nodes of the previous hidden layer.
	\item Hidden layer with 32 hidden nodes, using the rectifier  as nonlinear function for mapping the nodes of the previous hidden layer.
	\item Output layer for classifying the binary outcome (1=cartel, 0=competition), using the sigmoid function as nonlinear function for mapping the nodes of the last hidden layer.
\end{enumerate}
\end{figure}

We apply our approach of training and validating a CNN model to one part of our data, while we hold out another part for testing our model in unseen data (not used for the iterative training and validation steps) to check out-of-sample predictive performance, using accuracy (or the correct classification rate) as performance measure. For instance, when only considering Japanese data, we randomly choose 75\% of the images from the Okinawa region for the training and validation process and 25\% for testing performance. We follow the same approach when only investigating Swiss data or when analyzing a mixed (i.e.\ transnational) sample consisting of both Japanese and Swiss images. For each of the approaches considered, we repeat the procedure of randomly splitting the sample into training/validation and test data to develop and assess the predictive model 20 times. We report the mean, median, worst, and best performance across these 20 simulations, as the variance in performance is non-negligible across simulations due to the limited number of images available. Finally, we also use images from one country for training and validation to test performance based on images of the respective other country in order to check the transnational transferability of trained CNNs. We consider 20 simulations for this approach, too, even though the training/validation and test data are deterministic (in countries), while there is still a stochastic component within training and validation of the model.

\section{Empirical results}\label{empsec}

We first apply our CNN architecture outlined in Section \ref{cnn} to our 287 Japanese interaction graphs constructed from the Okinawa data \citep[see][]{Huberimhof2020}, running 20 simulations in which we randomly select 75\% of images for training and validation and the remaining 25\% for testing performance based on classification accuracy. Our outcome variable is a binary indicator taking the value 1 for a cartel period and 0 for competition. Table \ref{empir1} provides summary statistics for the accuracy in the total test data and separately for truly collusive or competitive test periods (i.e.\ the true positive and true negative rates) across simulations, namely the respective minimum, first quartile, median, mean, third quartile, and maximum. Despite the rather limited number of images available for training, the median and mean of accuracy amount to 91.6\% and 90.4\%, respectively, in the test data. This suggests that the CNN classifies on average roughly nine out of ten firms correctly as cartel members or competitive bidders. When investigating the accuracy within actually collusive and competitive cases, we find that the CNN performs somewhat better for detecting actual cartels, with an average and median of the true positive rate of 93.4\% and 92.6\%, respectively, than for detecting truly competitive firms, with a median and mean of the true negative rate of 89\% and 88.4\%, respectively. Put differently, we find median and mean classification error rates of 11\% and 11.6\%, respectively, under competition and of 8.4\% and 9.6\%, respectively, under collusion. Despite these slight imbalances across outcome classes, the CNN shows a quite decent overall predictive performance for both collusive and competitive cases. We, however, point out that the variation is non-negligible across our 20 simulations, as can be seen by the differences in the respective minimum and maximum accuracies. This suggests that one should ideally use a substantially larger pool of images for training and testing the model than was available to us.

%the true negative . They however declare a competitive firm as competitor with , respectively. The results thus suggest that CNNs perform less in detecting truly competitive firms than truly cartel participants, i.e. produce a bit more false positive than false negative results. We recall that false positive results are the error rate associated with competitive firms and is expressed as one minus the correct classification rate for competitive firms. In our case, the median and the mean of the error rate for competitive firms amount to 11\% and 11.6\%, respectively. The mean and the median of the error rate for the false negative results, i.e. for collusive firms not recognized as collusive, amount to 6.6\% and 7.4\%. Overall, the median and the mean of the overall error rate amount to 8.4\% and 9.6\%. Furthermore, three simulations out of four minimally deliver a correct prediction rate of 89.8\%. Such results imply that our detection method based on CNNs is able to correctly flag nine firms out of ten as either collusive or competitive.

{\renewcommand{\arraystretch}{1.1}
\begin{table} [!htp]
\caption{Summary statistics for accuracy when training and testing in the Japanese sample} \label{empir1}
\begin{center}
\begin{tabular}{lccccccc}\hline\hline
&Minimum&1st quartile &Median&Mean&3rd quartile&Maximum&Observations\\
All graphs&0.817&0.898&0.916&0.904&0.93&0.944&287\\
Collusion&0.771&0.906&0.934&0.926&0.957&1&143\\
Competition&0.658&0.862&0.89&0.884&0.923&0.951&144\\
\hline\hline
\end{tabular}
\end{center}
\par
%{\footnotesize Note: “All Graph.", “Coll. Graph" and “Comp. Graph" denote the accuracy for all the graphics, for the collusive graphics and for the competitive graphics, respectively. “Min", “Low.Q", “Median", “Mean", “Up.Q", “Max" and “N" denote for the accuracys the minimum, the lower quartile, the median, the mean, the upper quartile, the maximum for the accuracys in the 20 simulations, respectively. “N" indicates the number of observations}
\end{table}}

In a second step, apply the same procedure to our 240 Swiss interaction graphs and report the results in Table \ref{empir2}. Overall accuracy is again quite high, with its median and mean amounting to 91.7\% and 91.6\%, respectively. This is quite similar as for the Japanese data and suggests that our deep learning approach might work well in different countries. Contrary to the Japanese cases, however, the CNN now entails a larger true negative rate (of correctly classifying truly competitive firms) with a mean and median of 94.6\% and 93.6\%, respectively, while the respective numbers are 89.0\% and 88.4\% for the true positive rate (of correctly classifying truly competitive firms). Again, the overall performance appears very satisfactory (with roughly 9 out of 10 correct classifications) despite such imbalances across outcome classes of roughly 5 percentage points, but there is again some variation in the predictive performance across the 20 simulations.

{\renewcommand{\arraystretch}{1.1}
\begin{table} [!htp]
\caption{Summary statistics for accuracy when training and testing in the Swiss sample} \label{empir2}
\begin{center}
\begin{tabular}{lccccccc}\hline\hline
&Minimum&1st quartile &Median&Mean&3rd quartile&Maximum&Observations\\
All graphs&0.817&0.9&0.917&0.916&0.933&0.983&240\\
Collusion&0.769&0.862&0.875&0.884&0.917&1&96\\
Competition&0.784&0.918&0.946&0.936&0.973&1&144\\
\hline\hline
\end{tabular}
\end{center}
\par
%{\footnotesize Note: “All Graph.", “Coll. Graph" and “Comp. Graph" denote the accuracy for all the graphics, for the collusive graphics and for the competitive graphics, respectively. “Min", “Low.Q", “Median", “Mean", “Up.Q", “Max" and “N" denote the minimum, the lower quartile, the median, the mean, the upper quartile, the maximum for the accuracys in the 20 simulations, respectively. “N" indicates the number of observations}
\end{table}}

Next, we pool Japanese and Swiss graphs to obtain a mixed sample of 527 images in total and conduct the same simulation, training, validation, and testing steps as before. As reported in Table \ref{empir3}, median and mean accuracy amounts to 92\% and 91.4\%, respectively, which is rather similar as for the analyses within countries. However, the performance is now more balanced across outcome classes, as the means and medians of the true positive and true negative rates are rather similar. In addition, the variation of correct classification rates across the 20 simulations has generally decreased, an effect likely due to the larger number of images. All in all, these results appear very encouraging and suggest that well-performing CNNs might even be trained in mixed data composed of several countries or regions, which can be helpful for attaining larger samples.

{\renewcommand{\arraystretch}{1.1}
\begin{table} [!htp]
\caption{Summary statistics for accuracy when training and testing in both countries} \label{empir3}
\begin{center}
\begin{tabular}{lccccccc}\hline\hline
&Minimum&1st quartile &Median&Mean&3rd quartile&Maximum&Observations\\
All graphs&0.847&0.912&0.92&0.914&0.924&0.947&527\\
Collusion&0.776&0.901&0.925&0.916&0.951&0.969&239\\
Competition&0.84&0.902&0.912&0.913&0.932&0.969&288\\
\hline\hline
\end{tabular}
\end{center}
\par
%{\footnotesize Note: “All Graph.", “Coll. Graph" and “Comp. Graph" denote the accuracy for all the graphics, for the collusive graphics and for the competitive graphics, respectively. “Min", “Low.Q", “Median", “Mean", “Up.Q", “Max" and “N" denote the minimum, the lower quartile, the median, the mean, the upper quartile, the maximum for the accuracys in the 20 simulations, respectively. “N" indicates the number of observations}
\end{table}}

%So far our detection method based on CNNs using graphics depicting the interaction of one firm toward a group of firms exhibit outstanding classification rates for most of the simulations performed. It appears as a reliable detection method in two different countries using different bid-rigging cases and seems therefore to have a high potential for future application in different countries and to different cases.

We now turn to the more challenging task of training the predictive model in one country and testing it in the other country. As discussed in more detail in \citet{Huberimhof2020} for the case of Japan and Switzerland, the transfer of a trained method across borders may not yield the desired performance due to institutional differences in procurement, which may jeopardize the method's ability to distinguish collusion and competition in a country-specific context it has not been trained on. Table \ref{empir4} reports the summary statistics for accuracy when training based on Japanese graphs and testing based on Swiss graphs. While the Swiss test data are now identical across our 20 simulations, the Japanese images randomly chosen for training and validation generally differ. The median and mean accuracy attains 86\% and 85.4\%, respectively, is somewhat lower than before but still appears quite satisfactory. However, the imbalances across outcome classes are now more substantial. While the median and the mean of the true positive rate (among cartels in the Swiss test data) amount to 96.9\% and 96.1\%, respectively, the corresponding figures for the true negative rate (among competing firms in the Swiss test data) are 78.1\% and 78.3\%, respectively. Apparently, bidding interactions of Swiss firms under competition do not always yield a pattern that is close to that of Japanese firms under competition that were used for training the algorithm, while interactions look substantially more similar across Swiss and Japanese cartels. Yet, for neither outcome class, the predictive performance seems unacceptably low and overall accuracy is in fact quite good.

{\renewcommand{\arraystretch}{1.1}
\begin{table} [!htp]
\caption{Summary statistics for accuracy when training in Japan and testing in Switzerland} \label{empir4}
\begin{center}
\begin{tabular}{lccccccc}\hline\hline
&Minimum&1st quartile &Median&Mean&3rd quartile&Maximum&Observations\\
All graphs&0.775&0.841&0.86&0.854&0.872&0.892&527\\
Collusion&0.896&0.945&0.969&0.961&0.979&0.99&239\\
Competition&0.646&0.75&0.781&0.783&0.821&0.854&288\\
\hline\hline
\end{tabular}
\end{center}
\par
%{\footnotesize Note: “All Graph.", “Coll. Graph" and “Comp. Graph" denote the accuracy for all the graphics, for the collusive graphics and for the competitive graphics, respectively. “Min", “Low.Q", “Median", “Mean", “Up.Q", “Max" and “N" denote the minimum, the lower quartile, the median, the mean, the upper quartile, the maximum for the accuracys in the 20 simulations, respectively. “N" indicates the number of observations}
\end{table}}

Finally, we swap the roles of the countries and train the CNN based on Swiss graphs to test it based on Japanese graphs. As shown in Table \ref{empir5}, accuracy in all of the test sample is now lower with a median and mean of 79.4\% and 78.8\%, respectively, but nevertheless appears acceptable. However (and even more so than in the previous analysis), there is a sizable imbalance in the performance across classes. While the correct negative rate is very high with its median and mean amounting to 95.1\% and 95.2\%, respectively, the correct positive rate is substantially lower, with a median and mean of just 64.7\% and 62.3\%, respectively. This suggests that several Japanese collusive graphs are hard to distinguish from Swiss competitive graphs used for training, while Japanese competitive graphs can be very clearly distinguished from Swiss collusive graphs. Overall, we conclude that the CNN architecture yields a very promising predictive performance for classifying collusive and competitive bidding interactions of firms. Aside from a lower accuracy for specific outcome classes when only training in one country to test in the other one, the deep learning approach performs generally very well (with accuracies typically around 90\% or slightly higher) when training and testing within country-specific data or a mixed sample consisting of pooled graphs from both Japan and Switzerland.

{\renewcommand{\arraystretch}{1.1}
\begin{table} [!htp]
\caption{Summary statistics for accuracy when training in Switzerland and testing in Japan} \label{empir5}
\begin{center}
\begin{tabular}{lccccccc}\hline\hline
&Minimum&1st quartile &Median&Mean&3rd quartile&Maximum&Observations\\
All graphs&0.697&0.777&0.794&0.788&0.813&0.85&527\\
Collusion&0.413&0.612&0.647&0.623&0.678&0.755&239\\
Competition&0.924&0.943&0.951&0.952&0.958&0.986&288\\
\hline\hline
\end{tabular}
\end{center}
\par
%{\footnotesize Note: “All Graph.", “Coll. Graph" and “Comp. Graph" denote the accuracy for all the graphics, for the collusive graphics and for the competitive graphics, respectively. “Min", “Low.Q", “Median", “Mean", “Up.Q", “Max" and “N" denote the minimum, the lower quartile, the median, the mean, the upper quartile, the maximum for the accuracys in the 20 simulations, respectively. “N" indicates the number of observations}
\end{table}}

\section{Conclusion}\label{con}

This paper contributes to the literature on data-driven detection of bid-rigging cartels by proposing an novel method based on a deep learning architecture involving convolutional neural networks (CNN) for image recognition. We to this end construct graphs on pairwise bidding interactions of firms across various tenders, as also considered in the bid rotation test suggested in \citet{Imhof2018} and \citet{Imhof2019}, and use them as inputs for the CNN. We obtain an average out-of-sample accuracy of roughly 90\% or higher in classifying graphs with collusive and competitive interactions when applying the method separately to Japanese and Swiss data on bid-rigging cartels. Pooling the graphs from both countries even yields a slightly higher accuracy, which is also more balanced across outcome classes, i.e.\ truly collusive and competitive cases. Only when training (or developing) CNN models based on the graphs of one country in order to test the predictive performance in the respective other country, accuracy generally decreases. Yet, we still obtain an average accuracy of 85\% when training in the Japanese and testing in the Swiss data and of 79\% when training in the Swiss and testing in the Japanese data. However, imbalances in accuracy across collusive and competitive cases increase importantly when transferring trained models to another country, implying that the method no longer works similarly well for correctly predicting either outcome class. Nevertheless, the fact that we attain a high predictive performance in most scenarios considered despite our limited sample of just a few 100 graphs suggests that our CNN approach for bidding interaction classification is a very promising tool for screening procurement markets by inferring if some firm is likely a cartel member or not.

%If CNN models correctly predict Japanese competitive graphics and Swiss collusive graphics with accuracys above 95\%, they are less reliable in predicting Japanese collusive graphics and Swiss competitive graphics. The bidding pattern for some Japanese collusive and Swiss competitive graphics seems not to be sufficiently different for obtaining a higher accuracy. Considering that, Swiss and Japanese graphics are produced in different situations and different institutional environments, accuracys of 85-86\% and 79\% are encouraging results despite of an increased imbalance when training CNN models in one country and testing them in another.

In line with the impressive successes of deep learning in many domains like natural language processing and pattern recognition, our approach has the potential to outperform other quantitative approaches for flagging bid rigging and is one of the rare applications of CNNs in economics to date. Since our method focuses on the bidding interactions of a reference firm with other firms participating in the same tender, it (in contrast to most other approaches) permits identifying potential cartel members directly for further scrutiny. Indeed, a similarly high accuracy as in our analysis, implying that the CNN correctly classifies nine out of ten firms as cartel members or competing bidders, provides a solid base for competition agencies for the decision to open an investigation against a suspicious firm. Moreover, our method is also well suited for ex-post screening purposes. That is, after a competition agency has identified a group of conspicuous contracts, it can use our method to learn if firms participating in those conspicuous tenders exhibit a different interaction pattern than those firms not bidding in such tenders. If this was the case, the trained CNN model should be able to distinguish the bidding pattern of the conspicuous firms from the pattern of the non-conspicuous firms in a test sample. %If in addition data from previous cases are available, a competition agency can in a second step verify if the bidding pattern associated to the conspicuous firms is similar to the bidding pattern observed in previous bid-rigging conspiracies.

Even though the scope for applications of CNNs appears ample for fighting collusion and fraud more generally, a practical issue is that there is an infinite number of ways a CNN could be constructed in terms of tuning parameters like the number/size of filters (for extracting patterns), hidden layers (determining model complexity), nodes (nonlinear functions per layer), and many others. The best performing CNN architecture is a priori not obvious and needs to be iteratively explored, making deep learning seemingly more an art than a science. While the model analyzed in this paper yielded a very decent accuracy in the light of our comparably small data, it may well be that different specifications turn out to yield a higher performance in other contexts and when having available a larger amount of graphs on bidding interactions.

%search for the optimal predictive model structure is at the core of this research, our preliminary attempts with rather simplistic CNN models already suggest a high predictive performance when compared to previous studies in the literature. The detection method that we suggest is then reliable and seems to be very promising for future developments.

%Since our method focuses on the interaction of one firm toward a group of firms, we are able to directly identify potential cartel participants for further scrutiny. In fact, the quite high classification rates correctly flagging nine firms out of ten as either cartel participants or competitive firms offer a good confidence for competition agencies when deciding to open an investigation against a peculiar firm. Moreover, our test is also well designated for screening purpose. When a competition agency has identified a group of conspicuous contracts, it can verify with our detection method if the firms participating in those conspicuous contracts exhibit a significant different interaction pattern with the firms not bidding in conspicuous contracts. In this case, the predictive model based on CNNs will recognize with high classification rates that the bidding pattern of the conspicuous firms does not match the bidding pattern of the non-conspicuous firms. If data from previous cases are available, a competition agency can in a second step verify if the bidding pattern associated to the conspicuous firms is similar to the bidding pattern observed in previous bid-rigging conspiracies.

%\section*{Appendix A}

\pagebreak
\begin{spacing}{1.0}
\bibliographystyle{agu}
\bibliography {bibliothesis}
\end{spacing}

\end{document}